# MenakBERT - Hebrew Diacriticizer

Ido Cohen, Jacob Gidron, Idan Pinto
Tel Aviv University
{its.ido, jacob.o.gidron, idan.pinto3}@gmail.com,

## Abstract

Diacritical marks in the Hebrew language give words their vocalized form. The task of adding diacritical marks to plain Hebrew text is still dominated by a system that relies heavily on human-curated resources. Recent models trained on diacritized Hebrew texts still present a gap in performance. We use a recently developed char-based PLM to narrowly bridge this gap. Presenting MenakBERT, a character level transformer pretrained on Hebrew text and fine-tuned to produce diacritical marks for Hebrew sentences. We continue to show how fine-tuning a model for diacritizing transfers to a task such as part of speech tagging.

## 1 Introduction

Semitic languages use diacritical marks and vowels to give words their vocalized form. The ability to properly diacritize, or *dot*, text relies on a considerable understanding of the language. Undotted Hebrew text contains vastly more ambiguities compared to English text. Many words can be both understood and pronounced differently based on their diacritics, as shown in Table 1. Nowadays, these marks are rarely used so readers are left to infer the meaning from the context, we try to teach a machine to do the same.

| **Diacritized** | **Vocalization** | **Meaning** |
|---|---|---|
| בְּצִלָּם | Be-tzi-lam | In their shade |
| בְּצֶלֶם | Be-tze-lem | In one's image |
| בְּצַלָּם | Btza-lam | Their onion |
| בַּצַּלָּם | Ba-tza-lam | (*i.e. depends*) On the photographer |

Table 1: four ways to diacritize the same base letters leading to four different pronunciations and meaning.

To compensate for the inherent ambiguity in plain Hebrew script, most modern texts use a variation called *Full script (Nikkud Maleh)* instead of the standard *Partial script (Nikkud Haser).* Full script uses the letters י and ו which function like vowels in places where it can aid pronunciation. Thus, when the diacritical marks are dropped, these letters are maintained and so is the vocalization. These two coexisting variations present a challenge when training since they have slightly different rules for dotting.

It's worth pointing out the use of diacritical marks in Hebrew in contrast to other semitic languages. Marks in Hebrew are used mostly to determine the vocalization of a letter (and the word), and only in one case is it used to determine between two forms of the same base character. Some pairs of marks have vocalization that are almost indistinguishable to the modern ear, their difference so subtle it was neglected with time (i.e. *kamatz* and *patah*). So even knowing the correct vocalization, it is in most cases hard to choose the correct mark, adding to the ambiguity. In Arabic, in contrast, many marks distinguish between different consonants that have the same base character, and the rest determine the vocalization. In the case of the latter, each mark has a distinct vocalization.

Hebrew poses another challenge since most of the text obtainable today is not dotted, making it a challenge to assemble a sizable dataset. Diacritical marks are not used in everyday writing, nor in formal forms of writing, such as official documents or newspapers. These marks are mostly used for educating, so while most can read dotted text very few can produce it. Religious text is mostly dotted but both uses some extra marks and uses a form very different to modern text.

A word's diacritical marks are strongly linked to the context of word in the sentence, and the POS

tag of the word. Furthermore, when observing the sequences of marks, it's easy to notice that they mostly follow certain patterns, usually linked to the conjugation or inflection of the word's base form. Luckily, patterns and relations are attributes neural networks capture well. This leads us to believe that a transformer model should perform well on this task, and that good dotting skills could be a good indication of a model's understanding of language, and thus, of its ability to perform better on other downstream tasks.

**Previous work:**

Early attempts at tackling Hebrew dotting or vowel restoration use a probabilistic approach, HMMs (Kontorovich, 2001; Gal, 2002). These achieved some success but had trouble generalizing to unseen data.

Jumping quite a bit forward, after neural networks showed promising results in many fields, some attempts to incorporate these is dotting systems were made. Dicta's Nakdan (Shmidman et al., 2020), argued that in this case machine learned models will benefit from the help of explicit knowledge of the language. Thus, they developed a complex system which includes a first stage of morphological analysis (using an LSTM), and a second stage which uses expert-conceived tables of dotting rules. This model is still today's SOTA. A more recent model, Nakdimon (Gershuni and Pinter, 2021) trained a Bi-LSTM to produce a dotted sequence directly from the given plain sentence, without preliminary stages or the use of external resources. While not surpassing the SOTA, it showed promising results, thus motivating us to further investigate this heading. By presenting a comprehensive test set and testing against the existing SOTA, Nakdimon provided a steppingstone for more to come.

In this work, we seek to continue this very line by harnessing a resource that is now almost elementary in every NLP task, pretrained language models. We present MenakBERT, a transformer-based model that accepts an undotted sequence of characters and produces a sequence of diacritical marks.

## 2 TavBERT – the model's backbone

TavBERT (Omri Keren et al., 2022) is a char-based Bert-style masked language model (Devlin et al. 2018). It was pretrained on the OSCAR corpus comprising of 3.1B Hebrew words, following the same technique as SpanBERT (Joshi et al. 2020). The hypothesis at the root of the work on TavBERT claimed that a character aware tokenization method is more appropriate when dealing with morphologically rich languages such as Hebrew, in comparison with the more common word-pieces approach. Their results ended up reasserting the potential of the more common sub word tokenization. TavBERT performed on par with existing word-pieces based models on morophological tasks. But performed relatively poor on tasks which require more semantic capabilities. Keeping these results in mind, we still believe that using a char-based PLM as a backbone for our model will show better results, for the following reasons. Adding diacritical marks is a task applied per-char and tokenizing every char independently allows more flexibility in inferring sub-word relations, which are as important in this case as inter-word relations. Also, morphological understanding is key in this task, and according to the paper, this seems the be TavBERT's strong side.

## 3 MenakBERT

Following Nakdimon's approach, we divide Hebrew diacritical marks to three categories: First (**S**) for the dot distinguishing two consonants sharing the same base character, shin (שׁ) vs sin (שׂ), it is used only with this letter. Second (**D**[1]) for the dot in the center of a letter that in some case changes pronunciation of certain letters, and in other cases creating a similar affect as an emphasis on the letter, or gemination. Third for the rest of the marks (**N**[2]), used mostly for vocalization.

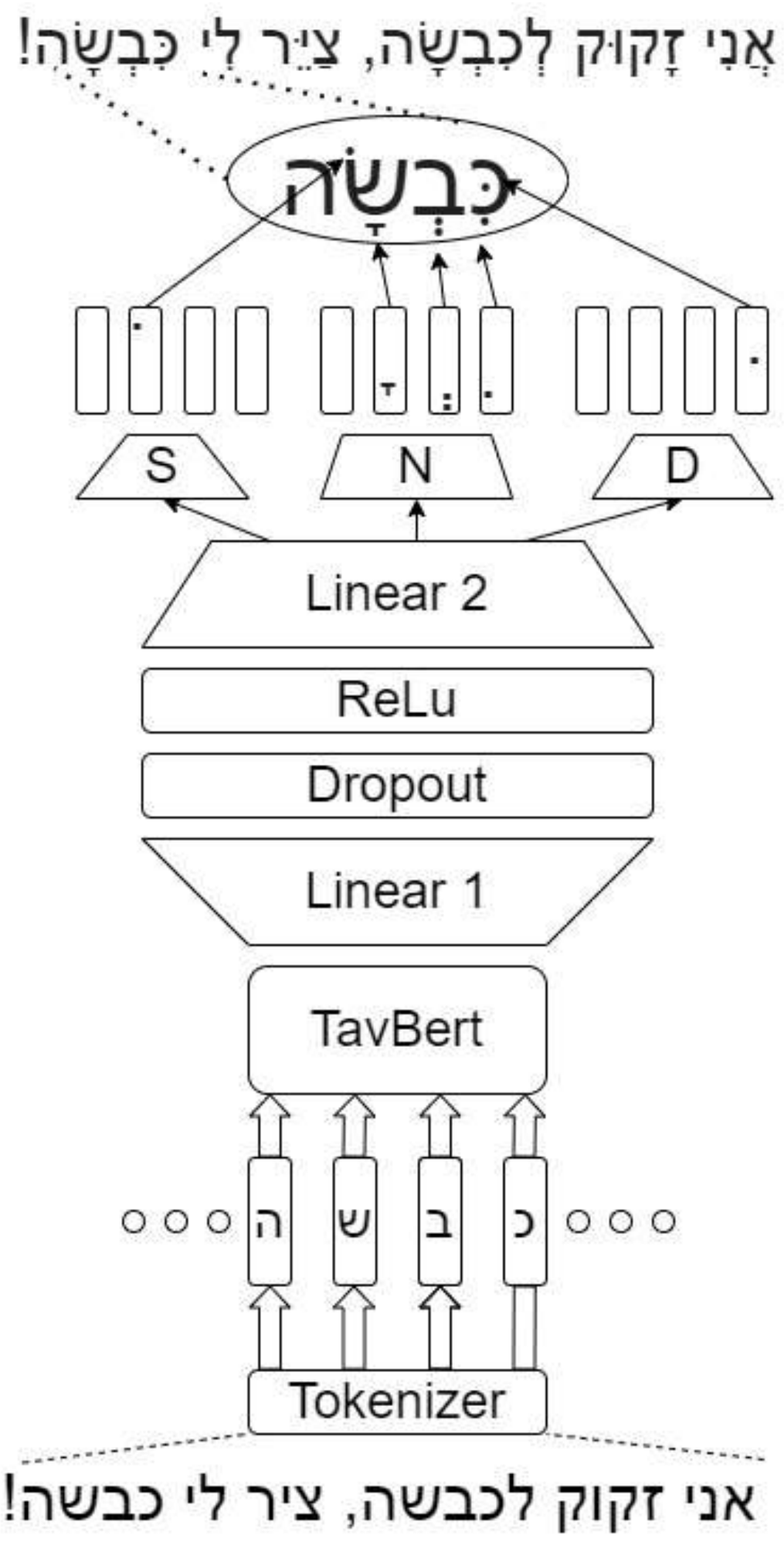

Figure 1: the entire MenakBERT pipeline.

**Character-transformer Dotter:**

As visualized in figure 1, MenakBERT takes as input a sequence of undotted characters and outputs three sequences of diacritical marks, for the three categories mentioned. It uses TavBERT's tokenizer to embed the input and pass it through the fine-tuned TavBERT transformer model. Next come two Linear layers, with a Dropout and activation layers in between. It then feeds the output to three distinct linear layers and SoftMax to produce three distributions for the three categories for every character. Taking the argmax of each, produces the three output sequences.

[1] short for Dagesh meaning emphasis in Hebrew.

[2] short for Niqqud which is the name for diacritical marks in Hebrew.

**Training data:**

Use cases for a dotting model would normally apply to modern text, both as an end goal (producing dotted text) and as a preliminary stage for other downstream tasks. For this reason, modern text is key in training such a model. Unfortunately, as mentioned before, dotted Hebrew text is hard to come by, especially modern. Nakdimon carried out impressive work creating a diverse dataset of both modern and premodern text. The premodern portion contains a diverse collection of texts from 12th century religious works to 19th century stories and poetry. The modern portion aims to reflect a more current form of expression. It contains modern books, songs, news and opinion articles, Wikipedia pages, judicial texts, and movie subtitles. Some of the modern texts were dotted using Dicta followed by manual error correction. The premodern dataset contains 2.6 million Hebrew characters, and the modern dataset contains 0.3 million Hebrew characters.

For the development set we use a dataset with only modern text. This will help steer the training in the right direction.

The dataset contains texts using both full script and partial script. We apply no preprocessing to change this and leave it to the model to understand the difference.

## 4 Experiments

In order to properly evaluate our model against earlier results we report the same metrics as Nakdimon: decision accuracy (DEC), which is computed over the entire set of individual possible decisions (for the three categories of diacritical marks). Character accuracy (CHA) is the portion of characters in the text that end up in their intended final form. Word accuracy (WOR) is the portion of words with no diacritization mistakes. Vocalization accuracy (VOC) is the portion of words where any dotting errors do not cause incorrect pronunciation.

| System | DEC | CHA | WOR | VOC |
|---|---|---|---|---|
| Snopi | 91.14 | 85.45 | 74.73 | 77.17 |
| Morfix | 97.25 | 95.35 | 89.43 | 91.64 |
| Dicta | **98.94** | **98.23** | **95.83** | **95.93** |
| Nakdimon | 97.37 | 95.41 | 87.21 | 89.32 |
| MenakBERT | 98.82 | 97.95 | 94.12 | 95.22 |

Table 2: results on the Nakdimon test set

We split the training data into four datasets: religious, premodern, early modern, and modern, and trained the model by chronological order so as to focus the model on the modern form. The texts are preprocessed in the same manner as in Nakdimon, for persistency reasons, with one difference. We choose to split the text to samples according to sentence boundaries, whereas Nakdimon splits by a set length. Optimization uses Adam with a learning rate scheduler with warmup. While training, we allow the gradient to backpropagate through the entire model, thus fine tuning the weights of the PLM TavBERT..

**Results**

The results shown in table 2 are on the test set created by Nakdimon. We compare our results against the results reported by Nakdimon of several systems. Dicta and Nakdimon are the two systems mentioned earlier. Morfix is a system developed in 1996 by Professor Jacob Shoike for the Center for Educational Technology. Snopy is a system developed by Tzvika Berkovich. We can clearly see two notable observations. The first is that MenakBERT closes most of the gap behind Dicta on all four metrics. The second is that in both WOR and VOC, where Nakdimon was unsuccessful at passing Morfix, MenakBERT passes Morfix and closes in on Dicta.

**Error analysis**

Unsurprisingly, in terms of types of mistakes, MenakBERT behaves similarly to Nakdimon. Not unlike other language models we see that MenakBERT learns patterns well. For example, conjugations of different verbs that share a pattern and often have the same diacritics. A pattern that, before dotting, could fit two parts of speech, could fool the model. For example, in this case the model dots the word as an adjective instead of a verb.

| Should be: | MenakBERT: |
|---|---|
| אִם לֹא תָּזוּזִי לֹא נַגִּיעַ בַּזְּמַן | אִם לֹא תְּזוּזִי לֹא נַגִּיעַ בַּזְּמַן |

In Hebrew the difference between a particular and a non-particular noun can be a hidden in the diacritics. In some cases, the model didn't choose the correct form, even when it succeeded with another noun in the same sentence. For example, in the sentence "Dealing with the money and property" MenakBERT marked: "Dealing with the money and a property". Or in Hebrew:

| Should be: | MenakBERT: |
|---|---|
| טִיפּוּל בַּכְּסָפִים וּבָרְכוּשׁ | טִיפּוּל בַּכְּסָפִים וּבְרְכוּשׁ |

English words and names are not uncommon in Hebrew text and reveal another of the model's weak spots. The model dots these words in a way that the vocalization has a Hebrew ring, but this will lead to a mistake. For example, a name like Sherill will be vocalized as Sharil.

**Qualitative evaluation**

We perform a qualitative evaluation by showing how the model reacts to certain patterns. We take a base form that does not exist in Hebrew (קלג) and build a sentence using different inflections of that form. We choose the inflections to fit the position and role in the sentence, this creates a sequence that sounds like a Hebrew sentence but complete gibberish. We bring here one example:

הַקֶּלֶג קִילֵּג אֶת הַקְּלָגוֹת בִּקְלִיגוֹת

Indeed, in all but one letter (the last ו) the model dots the words in the form that would fit a true word with the corresponding part of speech. This raises some interesting questions as to the importance of semantics when dotting. Whereas the importance of the syntactic structure is clear.

## 5 Development Experiments

### Dotting as masked prediction

As a first step before fine-tuning the model we tried testing if TavBert has any understanding of diacritics as a baseline. TavBert was trained as a masked model, so we modeled the problem as a masked prediction problem. When using UTF the diacritical marks appear as characters after the base character. TavBERT does not have a NULL character in its vocabulary, so it always outputs a character per mask. This makes it challenging to evaluate its performance, since allowing it to output marks wherever it decides by adding masks after every letter, forces the model to make mistakes. To address this, we give the model an easier challenge by replacing the real marks with masks. We do a qualitative evaluation, because a quotative one would yield no baseline for comparison.

When all marks were stripped from the sentence, the model behaved very poorly, and in most cases did not even attempt to output diacritical marks. Next, only marks from some words were removed, and replaced with masks. In this case the model did use diacritical marks to fill in the masks but with low success. One interesting observation is that when the word is a simple and common one that is not conjugated or inflected, the model usually gets it right! For example:

Should be:
לְאַחַר שֶׁמֶּחְקָרו הָרִאשׁוֹן פּוּרְסַם

Output:
לְאַחַר שֶׁמַּחָקְרוֹ הָרִאשׁוֹן פּוּרְסַם

In the above example the first two words' marks were masked. The first word, meaning *after*, was dotted correct. And the second, meaning *that his research*, was dotted wrong.

### Hyper parameter tuning

We experimented with the following hyperparameters using a coarse grid search followed by a fine one:

Dropout: Fiddling around with the dropout of the attention layers in the TavBERT portion of the model seemed to only hinder learning so we left it at 0.1 which was used in the TavBERT paper. The last dropout layer before the classification layers maxed at 0.3.

Last hidden layer size: We experimented with different sizes from 896 to 2048 and found that increasing the layer size helped, so we used 2048.

Learning rate: we used a learning rate scheduler with warmup. we set the warmup to 0.2 of the training steps. The initial learning rate was set to the default 0. We found a max learning rate of 1e-5 to give the best results.

Weights: Looking at initial results we realized that MenakBERT makes on average more mistakes on the second category *dagesh* (emphasis dot). We tried countering this using weighted cross-entropy. We scanned the entire training corpus to determine the distribution of the different classes and chose weights inverse to this distribution in order to focus the model on the rare marks. We tried this once again using the distribution only in the modern segment of the data. Both approaches showed no improvement and, in some cases, also interfered with training.

### Bible Bias

The bible provides a very large, dotted source. To our dismay, we noticed that when using it for training, we get worse results. We believe that since it becomes a great portion of the data, the model becomes biased towards certain patterns. And as we mentioned the bible doesn't conform to the modern dotting scheme. For this reason, we chose not to train on the bible. We bring one amusing example. In the sentence "in the ditch a whole platoon was under fire", the word 'whole' (*shlema)* is spelled the same as '(King) Solomon' (*shlomo)*, and even though the context is clear, the model still prefers dotting the word as Solomon.

### Sequential learning

Another observation addresses the order of training. We tried training over the entire training set as a whole vs. separating into four sets by chronological order and training sequentially, keeping the most modern text last. The second approach led to much better results when tested on modern texts. This notion reflects the idea of how the language evolves over time, letting the model learn from texts from all periods but focus on the most recent texts.

## 6 Dotting as prior for POS tagging

In Dicta's Nakdan, part of speech tagging was used as a first stage for diacritizing. We propose the opposite, to use knowledge in diacritizing for POS tagging. After fine-tuning MenakBERT for dotting on our dataset, we take out the weights of the fine-tuned TavBERT model and evaluate it against the original TavBERT model. We use the same method as in the TavBERT paper: We fine tune a token-classification head on top of each model's final encoder layer. The dataset used is the Hebrew treebank of the Universal Dependencies v2.2 dataset from the CoNLL-18 UD shared task (Sade et al., 2018).

**Results**

Figure 2 shows the accuracy per epoch for both the fine-tuned TavBERT and the original. We can see that the fine-tuned TavBERT enjoys a substantial head start in terms of accuracy, after only one epoch of training it scores 0.558 compared to the original 0.393. But a mere two epochs later the two models converge. After this point the two models even switch places for a while until completely converging. When observing F1 score, even the initial advantage vanishes.

It seems that Fine tuning the model for dotting did not provide the benefit we were hoping for.

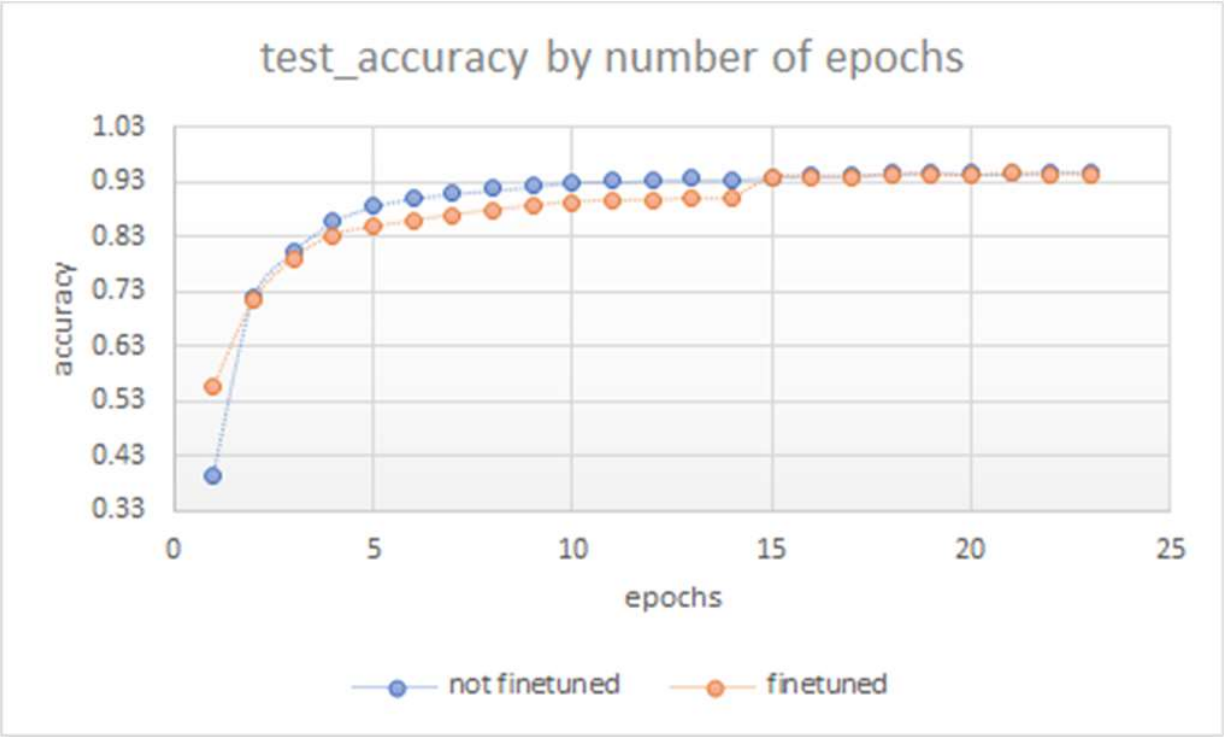

Figure 2

## 7 Conclusion

It is the general opinion amongst the NLP community that rule-based machines are 'out' and deep neural networks are 'in', but here we stumble across a case where the SOTA method for dotting text is still dominated by a hybrid technique using both an LSTM type architecture and a rule-based procedure. We believe that as in other fields of NLP this is temporary. Our results show that by using a PLM trained on Hebrew text we can narrowly close the gap in accuracy on Nakdimon's dataset. Furthermore, we think that this, still existing gap, is a testament to the scarcity of resources in Hebrew NLP. Specifically, in our case, a lack of modern diacritized text. This paper aims to be another piece in the growing foundations of the Hebrew NLP community and as further motivation to continue creating larger, more comprehensive and modern, datasets and language models.

## 8 Acknowledgments

We would like to thank Elazar Gershuni of Nakdimon for his priceless advice and for providing us with details about working with Nakdimon's model and dataset. We would like to thank Omri Keren for giving us access to his Hebrew PLM "TavBERT" which grew to be the backbone of MenakBERT.

## 9 References

Ya'akov Gal. 2002. An HMM approach to vowel restoration in Arabic and Hebrew. In Proceedings of the ACL-02 Workshop on Computational Approaches to Semitic Languages, Philadelphia, Pennsylvania, USA. Association for Computational Linguistics.

Leonid Kontorovich. 2001. Problems in semitic nlp: Hebrew vocalization using hmms.

Avi Shmidman, Shaltiel Shmidman, Moshe Koppel, and Yoav Goldberg. 2020. Nakdan: Professional Hebrew diacritizer. In Proceedings of the 58th Annual Meeting of the Association for Computational Linguistics: System Demonstrations, pages 197– 203, Online. Association for Computational Linguistics.

Elazar Gershuni, Yuval Pinter. 2021. Restoring Hebrew Diacritics Without a Dictionary (Nakdimon).

Omri Keren, Tal Avinari, Reut Tsarfaty, Omer Levy. 2022. Breaking Character: Are Subwords Good Enough for MRLs After All?

Jacob Devlin, Ming-Wei Chang, Kenton Lee, Kristina Toutanova. 2018. BERT: Pre-training of Deep Bidirectional Transformers for Language Understanding.

Mandar Joshi, Danqi Chen, Yinhan Liu, Daniel S. Weld, Luke Zettlemoyer, Omer Levy. 2020. SpanBERT: Improving Pre-training by Representing and Predicting Spans.

Shoval Sade, Amit Seker, and Reut Tsarfaty. 2018. The Hebrew Universal Dependency treebank: Past present and future

Amit Seker and Reut Tsarfaty. 2020. A pointer network architecture for joint morphological segmentation and tagging